\documentclass[runningheads]{llncs}

\usepackage[dvipsnames]{xcolor}
\usepackage{xspace}
\usepackage{amsmath}

\DeclareMathOperator*{\argmin}{arg\,min}
\newcommand{\method}[1]{{Time Reversal Fusion\xspace#1}}
\newcommand{\methodabbr}[1]{{TRF\xspace#1}}

\usepackage[dvipsnames]{xcolor}

\newcommand{\projectpage}{\href{https://time-reversal.github.io}{project page}}

\usepackage[mobile]{eccv}

\usepackage{eccvabbrv}

\usepackage{graphicx}
\usepackage{booktabs}

\usepackage{algorithm}
\usepackage{algpseudocode}
\usepackage{bm}

\algrenewcommand\algorithmicrequire{\textbf{Input:}}
\renewcommand*{\mathbf}[1]{\ifmmode\bm{#1}\else\textbf{#1}\fi}

\newcommand\blfootnote[1]{%
  \begingroup
  \renewcommand\thefootnote{}\footnote{#1}%
  \addtocounter{footnote}{-1}%
  \endgroup
}

\newlength\mylen

\usepackage[accsupp]{axessibility}  %

\usepackage[pagebackref,breaklinks,colorlinks,citecolor=eccvblue]{hyperref}

\usepackage{orcidlink}

\begin{document}

\definecolor{turquoise}{cmyk}{0.65,0,0.1,0.3}
\definecolor{purple}{rgb}{0.65,0,0.65}
\definecolor{dark_green}{rgb}{0, 0.5, 0}
\definecolor{orange}{rgb}{0.8, 0.6, 0.2}
\definecolor{red}{rgb}{0.8, 0.2, 0.2}
\definecolor{darkred}{rgb}{0.6, 0.1, 0.05}
\definecolor{blueish}{rgb}{0.0, 0.3, .6}
\definecolor{light_gray}{rgb}{0.7, 0.7, .7}
\definecolor{pink}{rgb}{1, 0, 1}
\definecolor{greyblue}{rgb}{0.25, 0.25, 1}

\title{Explorative Inbetweening of Time and Space}

\author{Haiwen Feng$^{\dagger}$\inst{1} \and
Zheng Ding$^{\dagger}$\inst{3} \and
Zhihao Xia\inst{2} \and
Simon Niklaus\inst{2} \and \\
Victoria Abrevaya\inst{1} \and
Michael J. Black\inst{1} \and
Xuaner Zhang\inst{2}
}

\authorrunning{Feng et al.}

\institute{
Max Planck Institute for Intelligent Systems
\and
Adobe
\and
University of California San Diego
\\
}

\maketitle

\begin{figure}
  \centering
  \vspace{-6mm}
  \includegraphics[width=\textwidth]{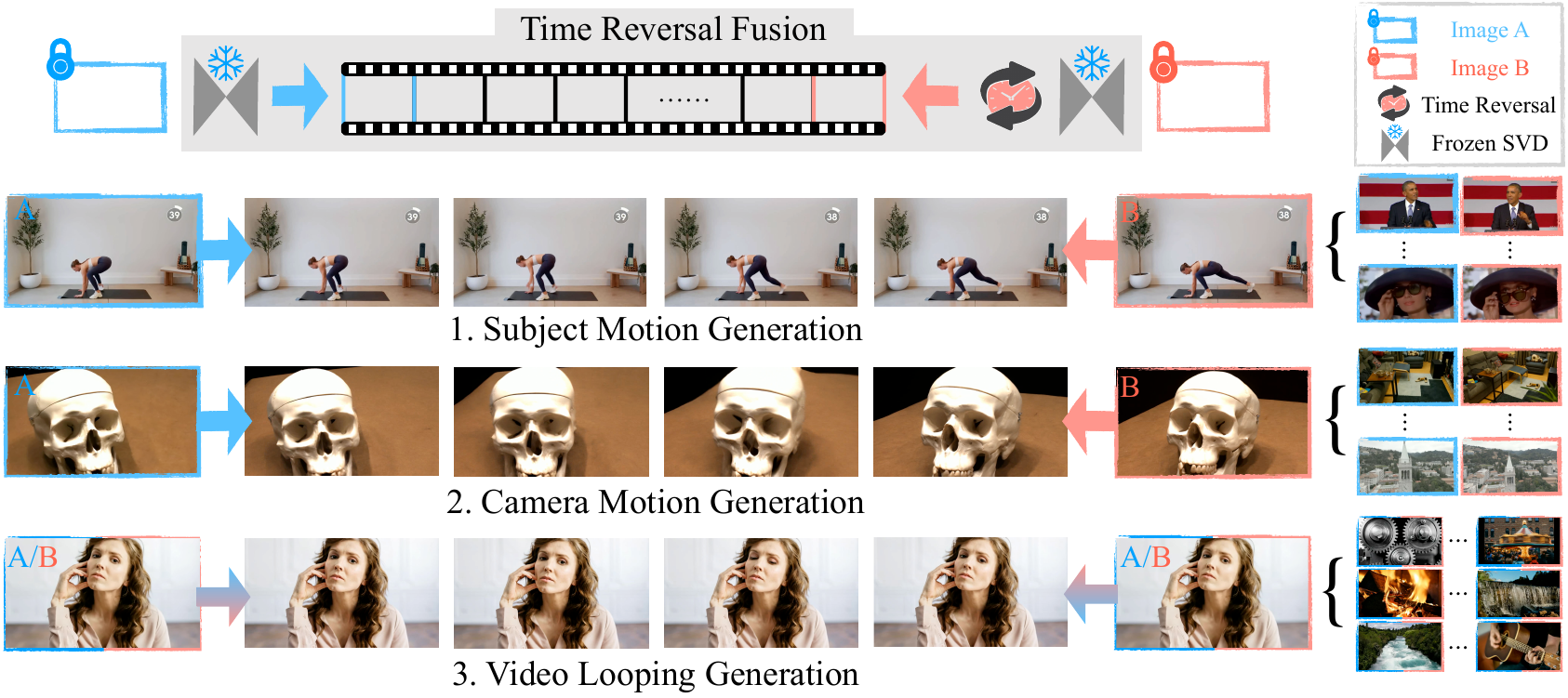}
  \caption{Bounded generation in three scenarios: 1) Generating subject motion with the two bound images capturing a moving subject. 2) Synthesizing camera motion using two images captured from different viewpoints of a static scene. 3) Achieving video looping by using the same image for both bounds. We propose a new sampling strategy, called Time Reversal Fusion, to preserve the inherent generalization of an image-to-video model while steering the video generation towards an exact ending frame.}
  \label{fig:teaser}
\end{figure}

\blfootnote{\scriptsize{$\dagger$ H.Feng partially performed the work and Z.Ding performed the work while interning at Adobe.}}

\vspace{-15mm}

\begin{abstract}
  We introduce bounded generation as a generalized task to control video generation to synthesize arbitrary camera and subject motion based only on a given start and end frame. Our objective is to fully leverage the inherent generalization capability of an image-to-video model without additional training or fine-tuning of the original model. This is achieved through the proposed new sampling strategy, which we call \method, that fuses the temporally forward and backward denoising paths conditioned on the start and end frame, respectively. The fused path results in a video that smoothly connects the two frames, generating inbetweening of faithful subject motion, novel views of static scenes, and seamless video looping when the two bounding frames are identical. We curate a diverse evaluation dataset of image pairs and compare against the closest existing methods. We find that \method outperforms related work on all subtasks, exhibiting the ability to generate complex motions and 3D-consistent views guided by bounded frames. See project page at \url{https://time-reversal.github.io}.
\end{abstract}

\section{Introduction}
\label{sec:intro}

The recent success of large image-to-video (I2V) models~\cite{blattmann2023stable, chen2023videocrafter1, bar2024lumiere} suggests that they have immense generalization capacity.  
These models can hallucinate complex dynamic scenes after exposure to millions of videos but they lack an important form of user control.
We often want to control generation between two image endpoints; that is, we want to generate the frames in between two image frames, which may be captured quite far apart in space or time.
We call this general task of inbetweening from sparse endpoint constraints \textit{bounded generation}.
Existing I2V models are incapable of doing bounded generation, since they lack the ability to control the motion towards an exact end state.
What we seek is a form of generalized control for video generation, capable of %
synthesizing both camera and object motion without making underlying assumptions on the motion trajectory.
For example, when the initial and final frames capture a dynamic subject, the task is to generate in-between object motion (Fig~\ref{fig:teaser} Row 1 shows complex articulated human motion). In instances where the snapshots capture a static scene from different viewpoints, the task is to fill in the camera trajectory (Fig~\ref{fig:teaser} Row 2 illustrates a rigid scene). If the start and end frames are identical, the task is to produce a looping video that starts and ends at the same frame (Fig~\ref{fig:teaser} Row 3 shows natural head, hand, and eye movements).
We define a general method to address all these problems in a unified manner.

On the surface, bounded generation is akin to several classic topics in the field, but with important differences: 
1) Frame interpolation aims to fill in content between frames by taking the shortest path along the arrow of time, whereas bounded generation exploits different plausible trajectories and can handle much larger frame gaps. 2) Novel view synthesis from wide-baseline stereo relies on generating new camera viewpoints through triangulation, necessitating the visibility of 3D points in both frames and the knowledge of camera poses, while bounded generation can generate novel views for any points present in either frame without any pose information. %
3) Single-image video looping hallucinates a flow field using specific motion models and requires scene segmentation, while bounded generation applies to arbitrary object motion without localizing any region.
These previous methods cannot solve the general bounded generation problem because they are constrained by inductive biases originating from either the domain-specific training data or the embedded physical model that addresses only specific types of motion. 
In short, they lack the capacity to generalize to arbitrary contexts.

In this paper, we bring bounded generation to I2V models by introducing a new sampling strategy: %
\textit{\method} (\methodabbr). 
\methodabbr is training- and tuning-free, thus it can harness the inherent generation capability of an I2V model.
We are motivated by empirical findings that existing I2V models are trained to generate content along the arrow of time, thereby lacking the ability to propagate image conditions backwards in time to preceding frames.
\methodabbr simultaneously denoises the temporally forward path conditioned on a given start frame and the backward path conditioned on an end frame, followed by the fusion of these two paths into a unified trajectory. 
We show that fusing the forward and backward paths can be achieved through an optimization objective, resulting in a straightforward averaging process.

Constraining both ends of the generated video make the problem challenging and naive approaches quickly become stuck in local minima, resulting in abrupt frame transitions.
To mitigate this, we introduce stochasticity through Noise Re-Injection to ensure smooth frame transitions.
\methodabbr combines bidirectional trajectories without relying on pixel correspondence or motion assumptions, resulting in video generation that predictably ends with the bounding frame.
Unlike existing controllable video generation methods~\cite{wang2023motionctrl, guo2023sparsectrl} that require training the control mechanism on curated datasets, 
our method does not require any training or fine-tuning, which allows it to fully leverage the original I2V model's generalization capacity. 

To evaluate videos created with bounded generation, we curate a dataset of 395 image pairs as the start and end bounds. These images contain snapshots ranging from multiview imagery of complex static scenes to kinematic motions of humans and animals, and also stochastic movement like fire and water. 
As our experiments show, bounded generation, when combined with large I2V models, not only opens up the possibility of numerous downstream tasks that were previously deemed hard, but also enables probing into the generated motion to understand the `mental dynamics' of I2V models. In summary, we propose:
\begin{enumerate}
    \item the task of \textit{bounded generation} for large image-to-video (I2V) models, where the goal is to synthesize the in-between frames given an arbitrary context by leveraging the generalization ability of these models.
    \item a novel sampling method that enables pretrained I2V models to perform bounded generation without fine-tuning or training.
    \item a dataset for bounded generation and a systematic evaluation of both our method and the closest existing work.  The empirical results indicates substantial improvements of our method over the state of the art. We will release the code and data to the research community for academic purposes. 
\end{enumerate}

\section{Related Works}

\subsection{Control-based Video Generation}
Diffusion-based video generation methods have recently achieved impressive results, with a focus on controllability—providing user-friendly ways to generate videos under controlled conditions. Initial efforts, inspired by the success of text-to-image models, concentrate on text-to-video generation \cite{ho2022imagen, ge2023preserve, luo2023videofusion, blattmann2023align, zhang2023show, mahapatra2023text, wang2023lavie}. Recognizing the limitations of text prompts in capturing complex scenes, later research \cite{guo2023animatediff, blattmann2023stable, li2023generative, li20233d} leverages image-conditioned video generation for a more direct approach. Notably, cinemagraph generation techniques \cite{holynski2021animating, mahapatra2023text, li2023generative, mahapatra2022controllable} focus on transforming still images into animated looping videos but are typically restricted to Eulerian motion, limiting their applicability to scenes with fluid continuous motion. Further innovations have introduced additional control mechanisms for video generation, such as structural guides \cite{esser2023structure, zhang2023controlvideo}, edge maps \cite{zhang2023controlvideo, khachatryan2023text2video}, and dynamic controls like motion trajectories \cite{wang2024videocomposer, yin2023dragnuwa, wang2023motionctrl}, camera poses \cite{wang2023motionctrl}, and sequences of human poses \cite{hu2023animateanyone}. Our work introduces a unique concept, \textit{bounded generation}, as a novel control mechanism for video generation, leveraging both start and end frames to guide the generation process. Using the same frame as the start and end guidance, our approach also enables the creation of looping videos, without relying on predefined motion models.

\subsection{Bounded Frame Generation}

Several existing sub-fields can be viewed as special cases of bounded frame generation. 
Our formulation unifies these, solving them with a unified framework that leverages a large video diffusion model.

\subsubsection{Frame Interpolation.}
There is an extensive history of research on frame interpolation, with early work focusing on finding heuristics for block-level motion compensation~\cite{choi2000new, ha2004motion}, while current research leverages machine learning instead~\cite{niklaus2017video, niklaus2017video, liu2017video}. Regardless of the underlying approach, video frame interpolation aims to find the most probable arrow of time that occurred between two frames. Looking at it differently, given two input frames it postulates that all motion follows the shortest path, which implies a single solution. This holds true even for techniques that aim to perform extreme versions of frame interpolation~\cite{reda2022film, sim2021xvfi}, or ones that take more than two input frames and then assume a quadratic path~\cite{xu2019quadratic, liu2020enhanced}. In contrast, our work focuses on ``explorative inbetweening'' where we are interested in the set of possible solutions that lead from one frame to another. Furthermore, we target inbetweening of distant inputs to increase the diversity of solutions. On this note, such dissimilar inputs go beyond the typical scenario that current frame interpolation techniques can handle.

\subsubsection{Sparse Novel View Synthesis.}
Recent advancements in novel view synthesis, sparked by the introduction of neural radiance fields (NeRF) \cite{mildenhall2020nerf}, have made significant strides \cite{sitzmann2019scene, trevithick2021grf, kerbl20233d, muller2022instant, barron2022mip}. The core idea is to utilize correspondences across multiple images with small baseline separations to reconstruct 3D geometry and appearance for generating new views of the observed 3D points. Efforts have been made to achieve synthesis from very sparse observations \cite{yu2021pixelnerf, deng2022depth, niemeyer2022regnerf, zhou2023sparsefusion, gu2023nerfdiff}, often leveraging priors learned from large datasets, including image priors in diffusion models \cite{zhou2023sparsefusion, gu2023nerfdiff}.
Du et al.~\cite{du2023learning} introduce a method for rendering new views from wide-baseline stereo pairs by employing cross-attention to match epipolar features between two frames. However, this approach requires known camera intrinsics and extrinsics and struggles with occluded points not visible in both views.
In contrast, our method diverges fundamentally from existing approaches to novel view synthesis. We do not rely on explicit 3D geometry modeling or a rendering pipeline. Instead, we generate novel views, even for 3D points visible in only one view, by harnessing the capabilities of a video diffusion model to perform bounded generation from two views of a static scene without needing any information on camera poses.

\subsection{Sampling-based Guided Image Generation}
Adopting new sampling techniques for manipulating the generative process of pretrained diffusion models is effective in a range of controlled image generation tasks \cite{lugmayr2022repaint, meng2021sdedit, bar2023multidiffusion, choi2021ilvr, tumanyan2023plug, mokady2023null, hertz2022prompt}.
For instance, RePaint \cite{lugmayr2022repaint} integrates observed regions into the sampled area during denoising for inpainting. SDEdit \cite{meng2021sdedit} applies noise to a user's stroke-guided image, then denoises it using a pre-trained diffusion model. To create large content images, like panoramas, DiffCollage \cite{zhang2023diffcollage} and MultiDiffusion \cite{bar2023multidiffusion} use a pre-trained diffusion model to generate segments of the content in parallel, merging the outputs at each denoising step for a cohesive large-scale image.
Our approach aligns with these concepts but targets video generation. By running two parallel I2V generations guided by start and end frames, we merge the outputs of each denoising step. This produces a coherent video bounded by the initial and final frames, marking a unique application of manipulating the generative process for video creation.

\section{Method}
\label{sec:method}

The goal of this work is to enable training-free \emph{bounded generation} within a diffusion image-to-video (I2V) framework -- that is, the use of contextual information in the form of a start and an end frame. 
We focus in particular on Stable Video Diffusion~\cite{blattmann2023stable} (SVD) which has shown impressive realism and generalization capacity for \emph{unbounded} video generation. 
While bounded generation can in principle be addressed by fine-tuning the model with paired data, 
this would inevitably lead to a compromise in model generalization~\cite{qiu2024controlling}.
Therefore, our study aims at training-free approaches.

We begin by reviewing SVD in Sec~\ref{ssec:method_alternatives}, and discuss two alternative and straightforward strategies for training-free bounded generation: condition manipulation and inpainting. We then analyze the reasons why these approaches are insufficient for our setting. Motivated by this, we present our proposed approach, Time Reversal Fusion, in Section~\ref{subsec:efg}. 

\subsection{Preliminaries}
\label{ssec:method_alternatives}

\paragraph{Stable Video Diffusion (SVD)} has achieved state-of-the-art performance in image-to-video generation, producing high-fidelity video sequences. Given an initial input frame, SVD generates a sequence of $N$ video frames, denoted by $\mathbf{x} = \{x^0, x^1, ..., x^{N-1}\}$. This sequence is constructed through a denoising diffusion process where, at each denoising step $t$, a conditional 3D-UNet $\Phi$ is used to iteratively denoise the sequence: %
\begin{equation}
    \mathbf{x}_{t-1} = \Phi(\mathbf{x}_t, t, c).
\end{equation}
Here, $c$ represents the condition of the initial input frame, which includes its CLIP~\cite{radford2021clip} embedding as well as its VAE latent, and ensures a consistent reference to the original frame throughout the video generation process. Note that SVD operates within a latent diffusion framework, meaning that, at the conclusion of the denoising steps, each frame within $\mathbf{x_0}$ is subsequently decoded back to pixel space using a VAE decoder.

There are two straigthforward solutions for incorporating \emph{bounded} generation within SVD: (1) condition manipulation, and (2) temporal inpainting. We discuss each of these in the following, and elaborate on the reasons why these simple approaches do not work in our setting.

\begin{figure}[tb]
  \centering
  \includegraphics[width=\textwidth]{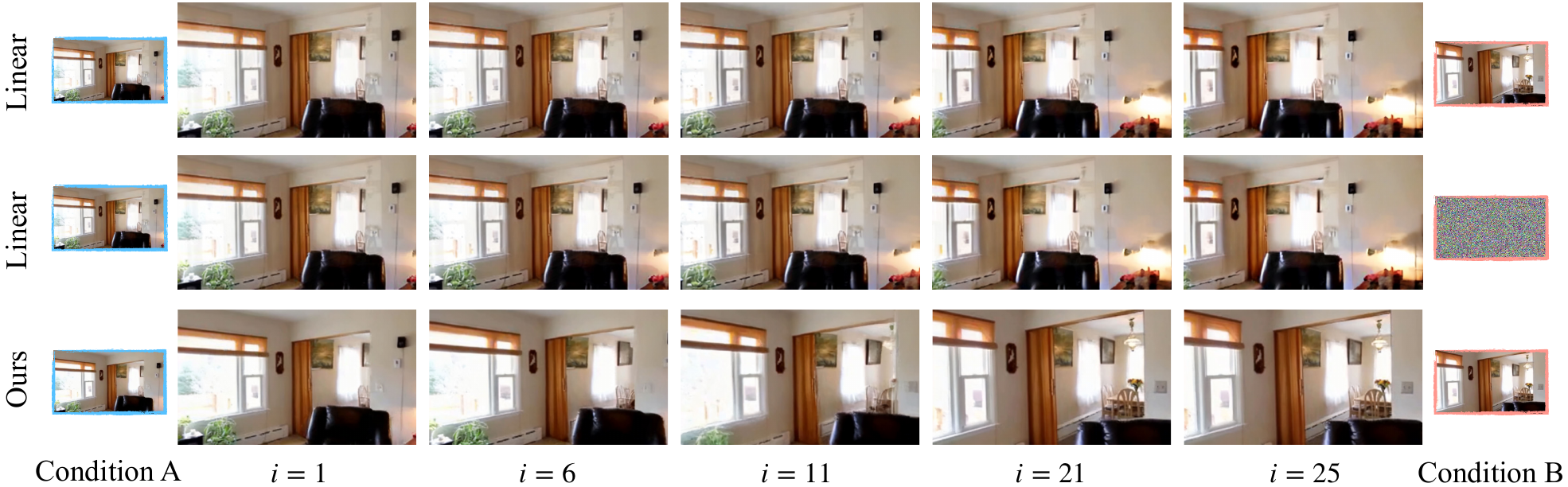}
  \caption{The impact of conditioning on video generation. We experiment with different conditioning strategy and show their effects on the generated video. (Row 1) Using a linear interpolation of A and B as the image condition, the generated video does not end at B. (Row 2) Swapping B with random noise yields similar results, indicating B imposes minimal influence on the generated contents. (Row 3) With the proposed time reversal fusion, our generated video starts with A and ends at B.}
  \label{fig:condition_ab}
\end{figure}

\paragraph{Condition manipulation. }
As mentioned, SVD conditions each frame-wise latent noise on the initial input frame. A straightforward solution to incorporate end-frame control is to condition the beginning of the sequence on the start frame, while the later part is conditioned on the end frame. This can be achieved by conditioning on a linear interpolation between the first and last frame, with the weight set as a function of time. We implemented this and observed that the video produced did not align with the condition set on the later frames; in other words, the condition specified during the later frames was largely ignored by the model. An example of this is shown in Fig.~\ref{fig:condition_ab}, where the top row is generated with the aforementioned strategy, while the middle row is generated by setting random noise as the end frame. In both cases we can observe a similar generation, suggesting that only the initial frames are responsible for the structure and dynamics of the output. 
We hypothesize that this is due to the nature of the training data, which was constructed to ensure significant disparity between the start and last frame. 
Due to this, and as observed in our own experiments, the network is trained to ignore the conditioning image on the latter frames, and focus instead on following the right dynamics based on the previous frames. Therefore, the intuitive idea of modulating towards the end frame by altering the condition is not a viable solution. 

\begin{figure}[tb]
  \centering
  \includegraphics[width=\textwidth]{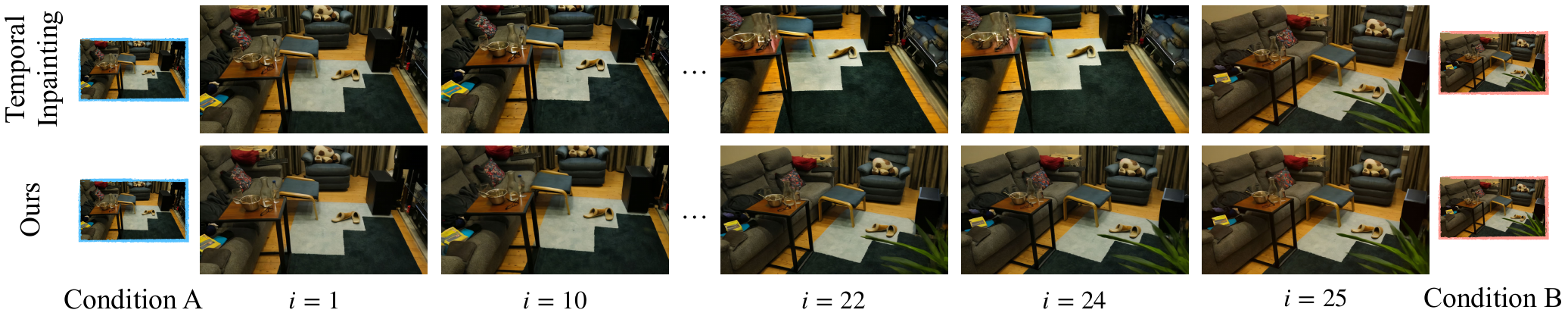}
  \caption{Image inpainting strategies do not apply to videos. We follow the standard diffusion inpainting method by replacing the last frame with the target frame at each denoising step. However, this results in a video that satisfies the end frame condition but with abrupt content changes, as indicated in the last frames in Row 1. Our method, on the other hand, generates a smooth video (Row 2) that ends at the given condition.}
  \label{fig:inpainting_ab}
  \vspace{-3mm}
\end{figure}

\paragraph{Temporal Inpainting. }
A second alternative for end-frame control is to apply diffusion image inpainting techniques~\cite{lugmayr2022repaint} to video data along the temporal axis. However, there are fundamental differences between videos and images that make these methods not applicable. 
First, images are static, and hence do not exhibit a preference for a direction, whereas videos are embedded with sequential influences that dictate a flow of time.
The architecture of SVD incorporates positional encoding of time stamps, imposing a temporal order to the generated video content. 
The learning process is designed to maintain temporal consistency, starting with a condition image (the first frame). As the sequence progresses, the later frames are trained to align more closely with the preceding frames rather than adapting to subsequent frames (as also analyzed in the paragraph above). 
In other words, each frame is temporally consistent primarily with its preceding frames, establishing an influence that follows the arrow of time. 
The reason why this hinders the application of an inpainting method is illustrated in the top row of Fig.~\ref{fig:inpainting_ab}. Here, the last frame is replaced with the target end-frame (with corresponding noise) at every denoising step, as done in e.g.~\cite{lugmayr2022repaint, meng2021sdedit}. This standard inpainting strategy leads however to a video where the end frame is correctly satisfied, but the rest of the generated frames do not naturally lead to it, resulting in abrupt changes.

To summarize, the unique characteristics of video data, coupled with the model architecture and learning patterns, highlight why techniques that are effective for images cannot be applied to videos. The directional bias, time-encoded architecture, and forward-leaning temporal consistency are integral to how the model processes and generates video content.
\vspace{-5mm}

\subsection{End-Frame Guidance using Time Reversal Fusion}
\label{subsec:efg}
Based on our analysis, we observe that SVD follows a \emph{forward} arrow of time, where the conditioning image initializes the video but its influence decreases over time. The challenge then lies in introducing a \emph{backward} influence to video generation, without fine-tuning the model. 

\begin{center}
\scalebox{0.8}{
\begin{minipage}{.55\textwidth}
\begin{algorithm}[H]
\caption{Method}\label{alg}
\newcommand{\ind}{\hspace{0mm}}
\begin{algorithmic}
\Require { \\
 $\mathbf{x}_T$: Random initialized noise; \\
   \ind $c_s$: Start frame conditions; \\
   \ind $c_e$: End frame conditions; \\
   \ind $M$: Number of noise injection steps; \\
   \ind $t_0$: Cutoff timestep for noise injection;\\
   \ind $\sigma_t$: Std at timestep $t$.}

\For {$t=T$..$1$}
  \State $\mathbf{x}_{t-1, s}=\Phi(\mathbf{x}_t, c_s, t)$
  \State $\mathbf{x}_{t-1, e}=\Phi(\mathbf{x}_t, c_e, t)$
  \State {$\mathbf{x}_{t-1}=\text{Fuse}(\mathbf{x}_{t-1, s}, \mathbf{x}_{t-1, e})\triangleright \text{Eq.}\ref{eq:fuse}$}
  
  \If {$t > t_0$}
    \For {$m=0..M-1$}
        \State {$\mathbf{\epsilon} \sim\mathcal{N}(0, \sqrt{\sigma_t^2 - \sigma_{t-1}^2} \mathbf{I})$}
        \State {$\mathbf{x_t} = \mathbf{x}_{t-1} + \mathbf{\epsilon}$}
        \State $\mathbf{x}_{t-1, s}=\Phi(\mathbf{x}_t, c_s, t)$
        \State $\mathbf{x}_{t-1, e}=\Phi(\mathbf{x}_t, c_e, t)$
        \State {$\mathbf{x}_{t-1}=\text{Fuse}(\mathbf{x}_{t-1, s}, \mathbf{x}_{t-1, e})\triangleright \text{Eq.}\ref{eq:fuse}$}
    \EndFor
  \EndIf
\EndFor
\end{algorithmic}
\end{algorithm}
\end{minipage}}%
\hspace{6mm}
\begin{minipage}{.45\textwidth}
\includegraphics[width=.9\linewidth]{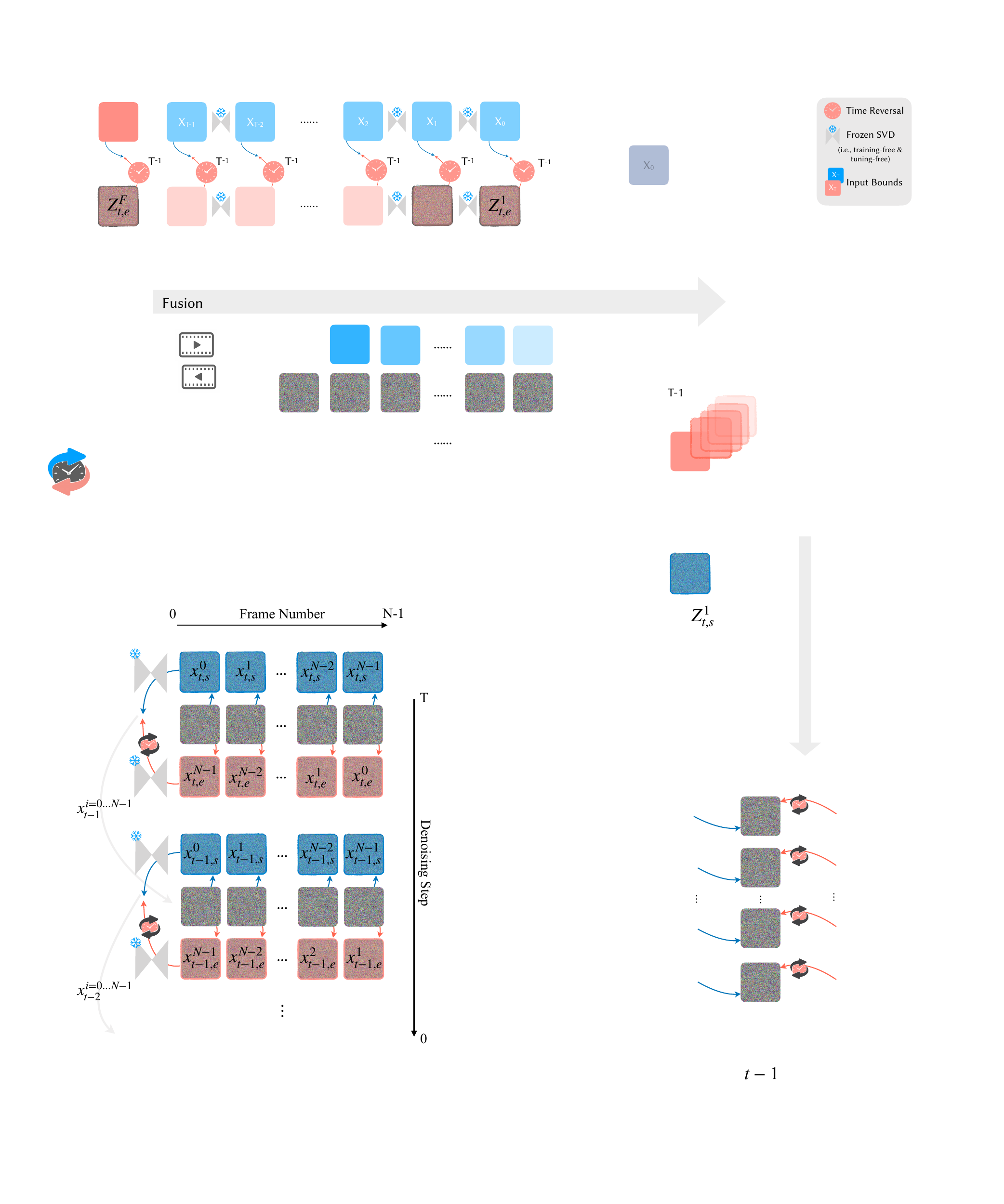}
\end{minipage}
\label{fig:method}
\captionof{figure}{Pseudo code and illustration of Time Reversal Fusion. Initiated with identical noise and conditioned on the start and end frame, the two paths undergo the SVD (frozen) denoiser. The forward path is fused with a time reversed backward path to produce the output for the subsequent step. Noise is re-injected to the fused output to add stochasticity in the sampling process.}
\end{center}

Our key idea is to generate two \emph{reference trajectories}: one conditioned on the starting frame \(c_s\), which we call forward generation, and another one conditioned on the ending frame \(c_e\), called the backward generation.
We initiate both forward and backward denoising paths from the same noise, conditioned on the corresponding frame. At each denoising step, the denoised output from the backward path is reversed such that its dynamics integrate better with the forward one. 
Drawing inspiration from large content generation works \cite{zhang2023diffcollage, bar2023multidiffusion}, we then composite both models into a single coherent video using a single sampling  strategy \(\Psi\), designed to align each denoising path closely with SVD's reference trajectories through the following optimization objective: 
\begin{equation}
    \small
    \Psi(\mathbf{x}_t, c_s, c_e) = \argmin_{\mathbf{x}} \left\| \mathbf{\alpha} \odot (\mathbf{x} - \Phi(\mathbf{x}_{t+1}, c_s)) \right\|^2 + \left\| (\mathbf{1-\alpha}) \odot (R(\mathbf{x}) - \Phi(R(\mathbf{x}_{t+1}), c_e)) \right\|^2.
\end{equation}
Here, \(R(\mathbf{x})\) represents the reverse of the sequence \(\mathbf{x}\), and \(\mathbf{\alpha} = \{\alpha_0, \alpha_1, ..., \alpha_{N-1}\}\) denotes a per-frame weighting factor, which is adjusted based on the proximity to the start or end guidance frame, either linearly or exponentially.

This optimization approach, a form of least squares approximation, naturally leads to a closed-form solution representing a weighted average of the forward and backward generations:
\begin{equation}
\label{eq:fuse}
    x_t^n = \alpha_n x_{t,s}^n + (1-\alpha_n) x_{t, e}^{N-n-1}.
\end{equation}
In this formula, \(x_{t,s}^n\) is the \(n\)-th frame from the SVD denoising UNet conditioned on the start frame, while \(x_{t,e}^n\) corresponds to the \(n\)-th frame conditioned on the end frame. This approach facilitates the generation of videos guided by initial and terminal frames through the nuanced interplay of forward and backward generation dynamics.

\subsubsection{Enhancing Fusion with Noise Re-Injection}

\begin{figure}[tb]
  \centering
  \includegraphics[width=\textwidth]{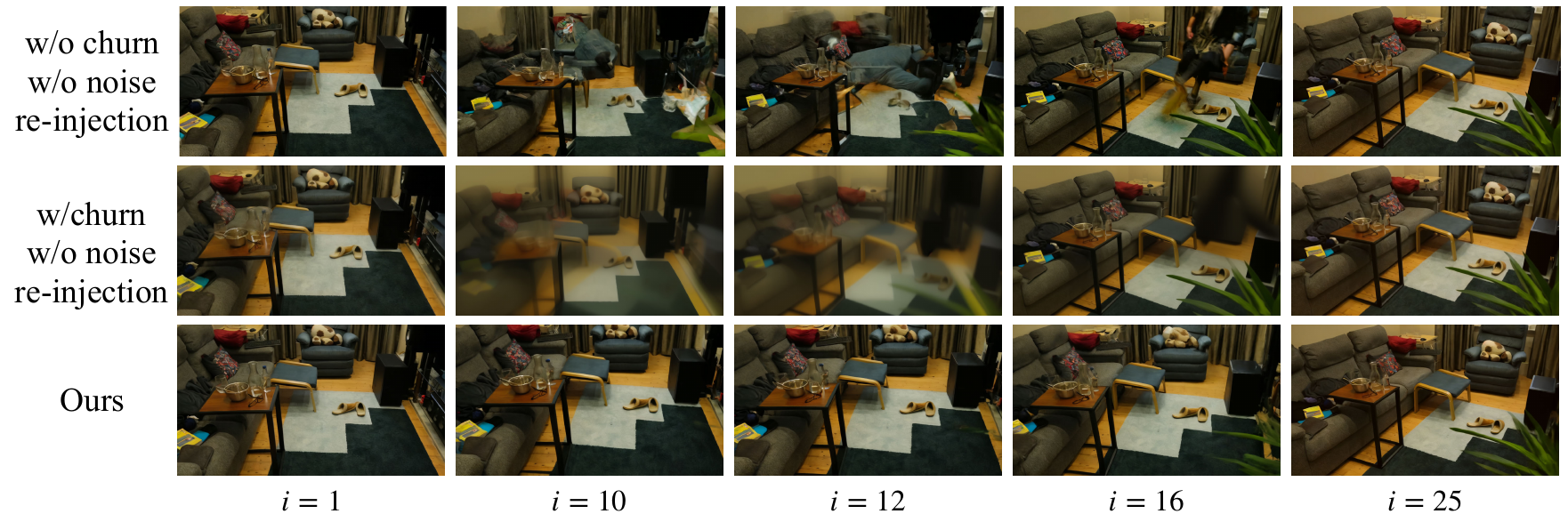}
  \caption{The impact of noise re-injection on fusion. (Row 1) Without any stochasticity, the video suffers from random dynamics and unsmooth transitions. (Row 2) Tuning the churn term in SVD leads to blurry and low-quality frames. (Row 3) Using noise re-injection leads to smooth and natural frame transitions.}
  \label{fig:noise}
  \vspace{-3mm}
\end{figure}

While time reversal fusion at each step effectively facilitates bounded generation, we occasionally observe blending cuts or undesirable artifacts, as highlighted in Fig.~\ref{fig:noise}, row 1. These issues often stem from significant disparities in the dynamics between the forward and backward generation processes. When such discrepancies are pronounced, the solution proposed in Eq.~\ref{eq:fuse} may result in poor quality, attributable to the lack of harmony between the two processes. Although subsequent denoising steps with SVD have the potential to enhance the quality, this improvement is typically short-lived, succumbing once again due to the same integration issue. Whereas the original denoising diffusion process ensures incremental quality enhancement with each step, the introduction of information and constraints from an alternate process can inadvertently alter the sampling trajectory.

To mitigate these discrepancies, we advocate for the introduction of additional stochasticity into the sampling process, thereby affording the network more opportunities to reconcile the two generative pathways. The EDM sampling strategy\cite{karras2022elucidating} employed during SVD inference incorporates a `churn' term to introduce noise at each step. We empirically find that it is not sufficient (Row 2 in Fig.~\ref{fig:noise}), as the small amount of noise introduced at each step is not strong enough to influence the generation at earlier denoising stage. To address this, we propose to augment each denoising step by injecting supplementary noise, followed by a denoising phase, and iterating this process several times before advancing to the subsequent denoising step. This approach of noise injection allows our sampling method to realign the fused generation at each step closer to the sampling trajectory defined by the pre-trained SVD, resulting in bounded generations that have similar visual fidelity as the SVD outputs. Our algorithmic approach is depicted in Fig.~\ref{fig:method}.

\section{Experiments}

We evaluate here our approach for bounded generation using the proposed Time Reversal Fusion (TRF). We consider three different scenarios for bounded generation: using dynamic bounds (different start and end frame with object/scene motion), view bounds (different start and end frame with camera motion), and identical bounds (using a single image as start and end frame). These scenarios are akin to the classic frame interpolation, novel view synthesis and loop video generation tasks but with more challenging constraints, and are detailed in Sec.~\ref{ssec:eval_setting}. In addition, we curate a new evaluation dataset for the aforementioned tasks containing difficult dynamics, which we present in Sec.~\ref{ssec:eval_dataset}. We compare against the closest state-of-the-art method for each of the tasks in Sec.~\ref{ssec:comparisons}, using standard metrics as well as a perceptual study, and show that our method significantly outperforms competitors. 

For more video results of all baseline comparisons, as well as additional video results of our method, please see our \projectpage.

\subsection{Evaluation Setting}
\label{ssec:eval_setting}
We categorize the test scenarios into the following three distinct settings, which cover diverse types of bounded generation: 

\noindent\textbf{Dynamic Bound:} With two frames capturing snapshots of a moving subject or object, the model should generate motion that seamlessly connects the frames. 

\noindent\textbf{View Bound:} When the two frames capture the same static object from different views, the model should synthesize plausible in-between camera trajectories, which also allows us to 
gauge the 3D consistency of the I2V model.

\noindent\textbf{Identical Bound:} When the two frames are identical, the model should generate looping videos that involve stochastic or periodic motion. 

\subsection{The Bounded Generation Dataset}
\label{ssec:eval_dataset}
To evaluate bounded generation on the three settings above, we curated a high resolution image / video dataset for each of the tasks, consisting of:

\noindent -- 115 image pairs sampled from YouTube videos, including kinematic motions of humans and animals, camera motion of complex scenes (e.g. landscape, cityscape, drone shots, etc.) and human-object interaction from movies, offering a broad spectrum of dynamic contents, paired with ground-truth clips.

\noindent -- 25 wide-baseline image pairs sampled from 6 indoor / yard scenes, plus 15 out-of-distribution image pairs ranging from underwater reef to crowded table, which goes beyond the typical room tour distribution. The image pairs are selected from existing novel view synthesis %
datasets~\cite{debevec1998image, mildenhall2020nerf, hedman2018deep, barron2022mip}.

\noindent -- 240 static images from pexels.com~\cite{pexels}, covering various dynamics such as natural phenomena (flaming, snowing), human activities (interaction with instruments, facial expressions), and larger scene dynamics (time-lapses, crowd movements) under 8 categories.

\begin{table}[tb]
  \centering
  \caption{Quantitative results on different downstream tasks: 1) dynamic bounds (Dyn.Bnd.) in terms of FVD~\cite{unterthiner2019fvd}, 2) identical bounds (Id.Bnd.) in terms of FVD, and 3) View bounds (View Bnd.) in terms of (a) FID with different feature dimensions, and (b) number of correspondences matched using COLMAP~\cite{schoenberger2016sfm, schoenberger2016mvs}.}
  \label{tab:combined}
  \begin{tabular}{@{}lc||lc||lccc@{}}
    \toprule
    Dyn.Bnd. & FVD$_{25}$ $\downarrow$ & Id.Bnd. &FVD$_{25}$ $\downarrow$ & View Bnd.  & FID$_{192}$ $\downarrow$ & FID$_{64}$ $\downarrow$ & COLMAP $\uparrow$\\    
    \hline
     FILM~\cite{reda2022film} & 656.88 & T2C~\cite{mahapatra2023text} & 911.67 & Du et al.\cite{du2023learning} & 28.70 & 8.67 & 379.61 \\
     Ours & \textbf{431.16} & Ours & \textbf{458.91} & Ours & \textbf{10.31} & \textbf{3.43} & \textbf{884.08} \\
    \bottomrule
  \end{tabular}
\end{table}

\vspace{-3mm}

\begin{figure}[tb]
  \centering
  \includegraphics[width=.9\textwidth]{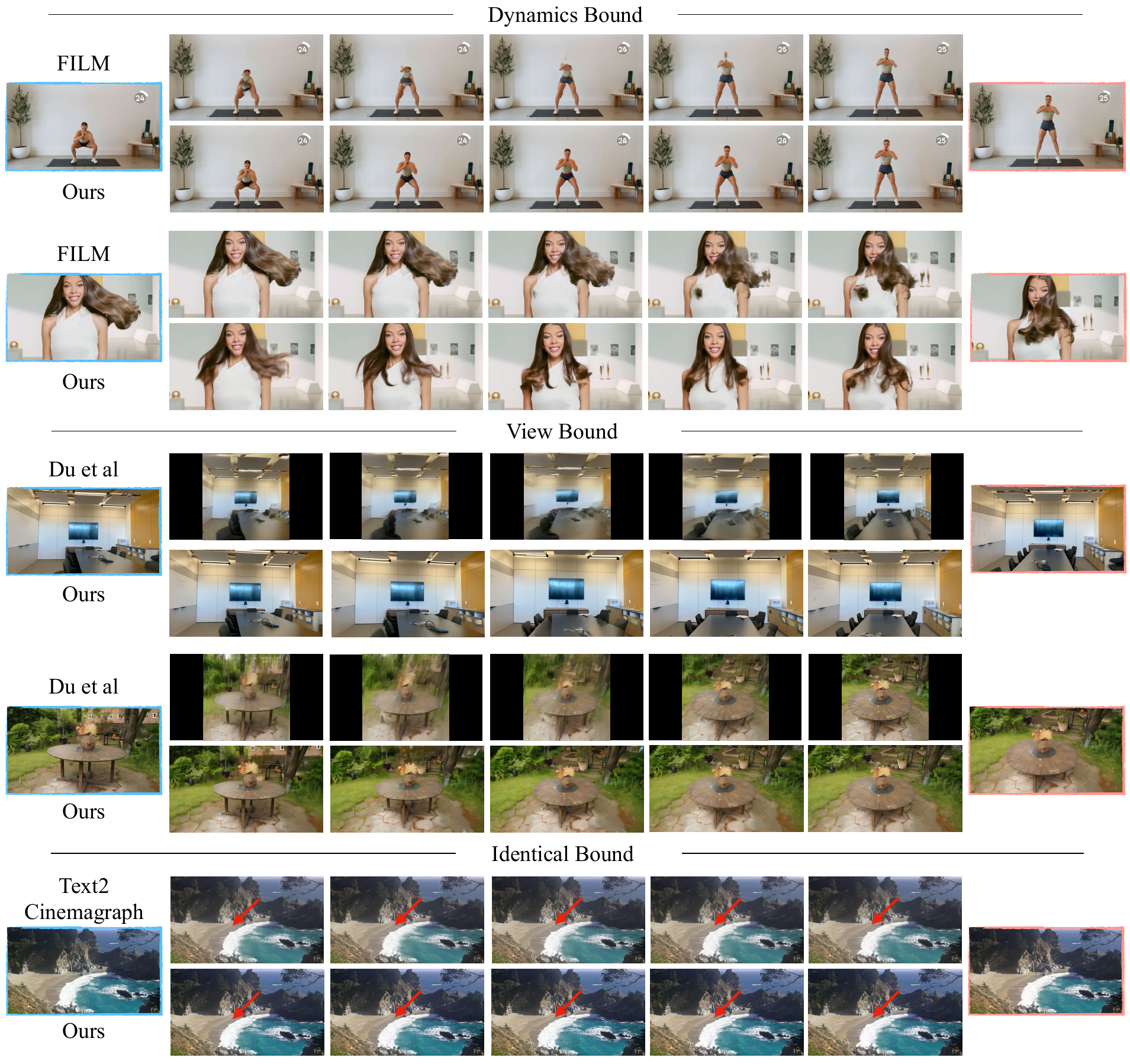}
  \caption{Baseline comparisons. While there is no existing work that simultaneously achieves the same goal across all three scenarios, we compare against the closest work within each category. For Dynamics Bound, FILM fails to interpolate between large or complex motions such as kinematics. For View Bound, Du et al shows artifacts of blurriness and stretching. 
  On Identical Bound, our generated looping video depicts a more natural movement of the wave. We suggest viewing the videos in the \projectpage.}
  \label{fig:baselines}
\end{figure}

\begin{figure}[tb]
  \centering
  \includegraphics[width=\textwidth]{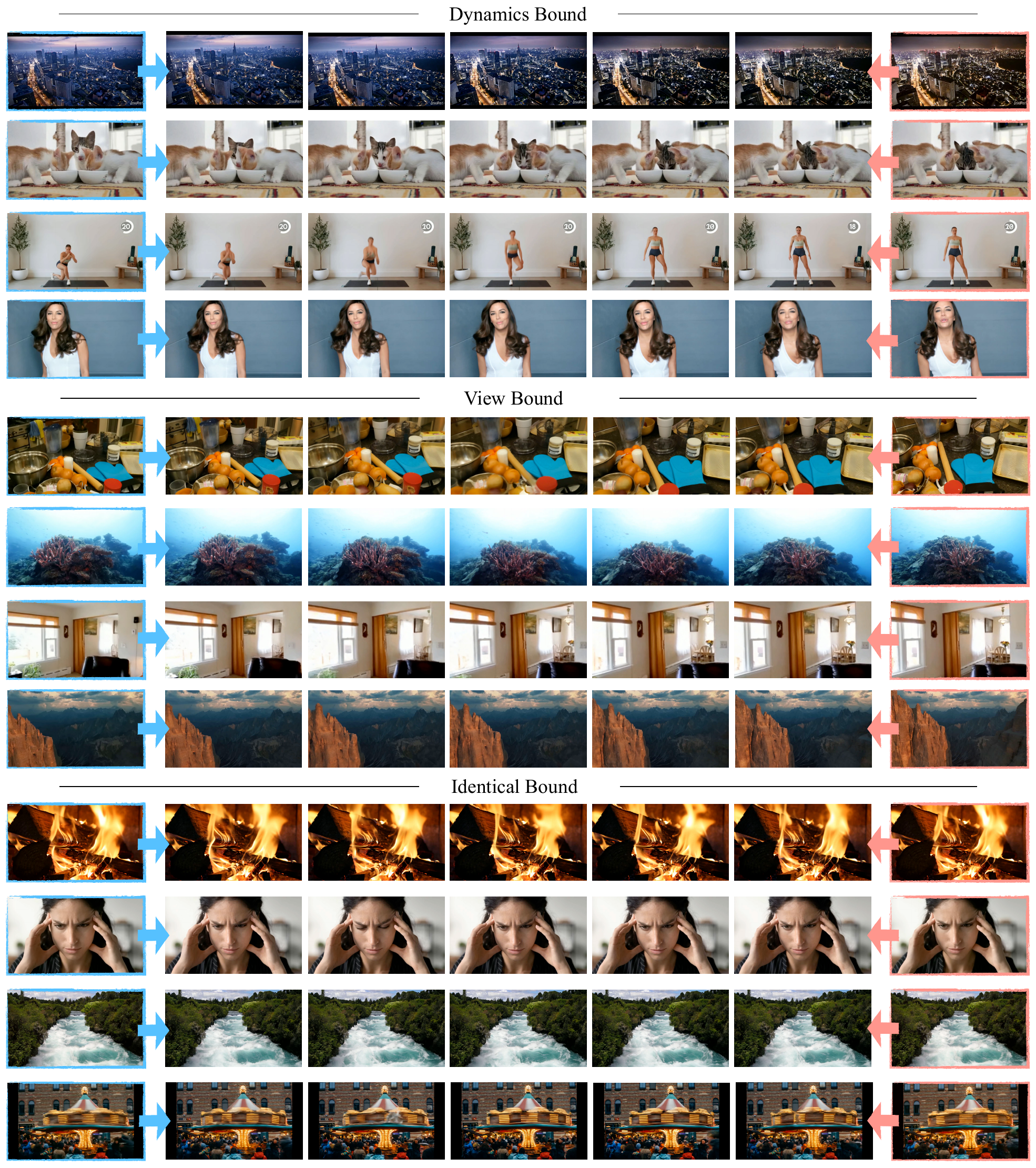}
  \caption{Additional results on the Bounded Generation Dataset. We show that by enabling bounded generation on an I2V model with TRF, we can leverage its great generalization capability to generate a diverse set of dynamics, including non-rigid body gestures, kinematics and nuanced hair movements. We can also synthesize camera trajectories across varied scenes and landscapes. The Identical Bound scenario produces looping videos capturing different movements in nature, which has not been shown before using a unified framework. Best viewed in video in the \projectpage.}
  \label{fig:our_result}
\end{figure}

\subsection{Comparative Analysis}
\label{ssec:comparisons}
We apply \methodabbr to a frozen image-to-video model, Stable Video Diffusion, to generate videos that are conditioned on the image pairs from our curated dataset.
While there is no existing method that accomplishes bounded generation across all three tasks, there are closely related works specific to each scenario. We compare against each state-of-the-art method, and demonstrate that they are not capable of fulfilling this new task.
Given that both task and method inherently involve hallucination, the use of pixel-aligned metrics for quality evaluation is impractical. Instead, we use FVD~\cite{unterthiner2019fvd} or FID~\cite{heusel2017gans} to compare the distribution between the generation and the ground truth. In the View Bound scenario, we further assess the 3D consistency of the generation based on the number of 3D corresponded points found by COLMAP, following the protocol in~\cite{cai2022diffdreamer}.

In the case of \textbf{Dynamics Bound}, the closest task is frame interpolation, which aims to smoothly interpolate between two given frames. We hence compare against FILM~\cite{reda2022film}, a large motion frame interpolation method, using FVD. As per the results displayed in %
Table~\ref{tab:combined}, 
TRF outperforms FILM by 30\%. This large performance gap is due to the fact that FILM cannot handle interpolation of motion that is too further apart, or complicated motion that requires semantic understanding. As shown in Fig.~\ref{fig:baselines} top, TRF is able to synthesize complex kinematics of human body and 3D consistent appearance in unseen regions, attributed to the underlying dynamic understanding and generalization ability of SVD, as well as the seamless dynamics trajectory fusing by TRF. In contrast, FILM primarily relies on flow-based correspondence and struggles to produce semantically meaningful motion when given sparse correspondences.

In the case of \textbf{View Bound}, the given frames are different viewpoints of the same static scene. We compare against the method Du et al.~\cite{du2023learning} that addresses wide-baseline novel view synthesis using neural rendering. Note that their model is trained with known camera poses or correspondences, while our method has access to neither. 
We use FID~\cite{heusel2017gans, Seitzer2020FID} to evaluate the quality of the rendered images in terms of fidelity to the ground truth, and evaluate 3D consistency by performing sparse reconstruction (SFM) from the generated novel views with COLMAP~\cite{schoenberger2016mvs, schoenberger2016sfm} following \cite{cai2022diffdreamer}. The number of extracted 3D correspondence points identified by COLMAP serves as an indicator of 3D consistency across all generated views. %
Given that Du et al. and SVD are trained under different resolution and aspect ratio, we resize and crop the ground-truth images to match their original size, and calculated FID accordingly. COLMAP is applied on both methods in the same area.
Quantitatively, TRF outperforms the baseline model by 60\% on FID, indicating the superior quality of our method. This is also verified by the qualitative results as shown in Fig.~\ref{fig:baselines} middle, where our rendered novel views contain less blurriness and stretching artifacts compared to the baseline, especially for 3D points that are only visible in one of the frames. 
The COLMAP reconstruction confirms that our rendered novel views are not only good in terms of visual quality, but also more 3D-consistent across the given wide-baseline paired views than the baseline method.

The task of \textbf{Identical Bound} is directly related to single-image cinemagraph. We hence compare against the recent work Text2Cinemagraph (T2C)~\cite{mahapatra2023text}, that uses images and texts to generate cinemagraphs with a dedicated pipeline including segmentation and motion prior training. We follow their evaluation protocol to compare FVD score on the validation set of Holynski et al.~\cite{holynski2021animating}. Table.~\ref{tab:combined} middle shows the substantial improvement of our method with around 50\% lower FVD score. The qualitative evaluation in Fig.~\ref{fig:baselines} bottom demonstrates that our results have more natural movement of wave, in contrast to the persistent wave of T2C, which often produces more subtle motions. We point the reader to our \projectpage{} for the video version of the results.

While T2C is specifically designed to animate the fluid motion of the segmented water region in the image, TRF can easily generalize to a larger range of motions, from stochastic dynamics of flame to non-rigid facial expressions of humans. This is thanks to the generative power of the pretrained I2V model, which enables generalization without any specific design choice or training data, as shown in Fig.~\ref{fig:our_result} bottom. Our Bounded Generation dataset (image dynamics subset) contains 12 different types of motion or interactions that qualitatively demonstrate TRF's generalization ability with identical bounds. More video results of diverse motion types can be found on the \projectpage{}.

\vspace{-3mm}

\subsubsection{Perceptual Study}
\label{ssec:user_study}

We also conducted a perceptual study to measure human preference between our method and the corresponding baseline.
Using Amazon Mechanical Turk (AMT), each participant was presented with 30 pairwise results. The participants were instructed to select the video they found more ``realistic, of higher quality, and exhibiting more natural motions and transitions''. In each pair, one video %
was randomly assigned to be from our method, while the other one was the corresponding generation from the closest baseline. The videos presented were randomly selected from either of the three tasks. 
To ensure the validity of the responses, we included 5 control trials within these comparisons with clearly unnatural videos. 
From this study, we collected 66 valid responses. The preference rate, indicating the proportion of participants favoring our method over the baseline, was then calculated based on the valid responses.

The results are shown in Table~\ref{tab:user_study}. The study shows a clear preference for our method in all three tasks with an overall average preference rate of 83.67\%. Particularly, we obtain the higher rate on view-bound results with a $97.79\%$ preference rate. Note that this task (generating camera trajectories from two sparse and unposed views) has traditionally been considered difficult, as also acknowledged by Du~\etal~\cite{du2023learning}. While the quality of their method significantly degrades when no camera pose is given, exhibiting blurry and unclear images, our work retains the sharpness and quality of SVD and generates reasonable camera trajectories. 

\begin{table}[h!]
    \centering
    \begin{tabular}{@{}lccc@{}}
    \toprule
        \hline
        Overall Avg.  & View bound  & Identical bound   & Dynamic bound  \\ \hline
        83.67\%   & 97.79\%     & 70.28\%          & 82.94\%         \\ \hline
        \bottomrule
    \end{tabular}
    \vspace{3mm}
    \caption{Perceptual study: Preference rates for each of the three subtasks, compared against the corresponding baseline (Du \etal~\cite{du2023learning}, Text2Cinemagraph~\cite{mahapatra2023text} and FILM~\cite{reda2022film}).}
    \label{tab:user_study}
\end{table}

\section{Discussion}
\label{sec:discussion}

\paragraph{Probing I2V models.} The bounded generation task along with TRF can offer a unique lens to assess SVD's world dynamics understanding. Given two observations, we can assess how the I2V model connects the motion trajectory, allowing us to compare the generated and the observed real-world dynamics. For example, the results on Dynamics Bound on the top of Fig.~\ref{fig:our_result} indicate the model's ability to understand and generate complex kinematics trajectories of articulated human bodies under different clothing, lighting, or with different image quality. Beyond articulated motion, the results of rows 2 and 4 indicate an ability to synthesize non-rigid motions like expression transitions and hair movements. In addition, the View Bound scenario exhibits 3D consistency across diverse real-world scenes, showcasing the model's generalization ability and 3D understanding of the physical world. The looping videos generated with identical bound indicate how well the model understands the implicit movement tendencies within a static image. These results suggest that applying similar techniques to other I2V models can serve as a way to probe the type and complexity of the dynamics that the model has learned. 

\paragraph{The importance of the motion bucket ID.} While our Time Reversal Fusion (TRF) method successfully achieves bounded generation without additional training, it does require careful tuning of the temporal conditioning parameters, such as motion bucket ID and frames per second (fps), to produce visually coherent outputs for different inputs. A critical aspect to note is the necessity for a match between the image content and the motion ID. This requirement stems from the underlying principles of Stable Video Diffusion (SVD), where the motion ID influences the intensity of pixel movement in the generated video – higher values result in more dynamic pixel behavior and vice versa. Selecting an appropriate motion ID range is crucial for each input image based on its dynamic contents; otherwise, the generated video may exhibit artifacts.
Interestingly, even though bounded generation poses a more complex challenge than straightforward sampling from SVD – requiring the model to generate specific motion trajectories that may not align with its typical motion distribution – our TRF method can effectively alleviate motion incompatibility artifacts. %
We believe this is due to the fact that the second view acts effectively as a constraint, providing additional guidance for the generation process. 
Through this we can mitigate the problem 
of motion ID in SVD, except in cases where the original motion ID is significantly inaccurate.
For example, in a static scene, a large motion ID may lead to excessive camera motion or unnatural addition of moving objects into the scene. Conversely, a smaller ID typically results in more subtle camera movements. However, if two wide-baseline views are significantly different, fusing them might inevitably lead to cut or blend effects due to insufficient dynamics that can seamlessly bridge the views.

\paragraph{Limitations.} One limitation of our method stems from the stochasticity involved in the generation of the forward and backward passes. For two given images, the distribution of motion paths that SVD can take might vary significantly. This means that the start- and end-frame paths could generate very different videos, resulting in an unrealistically fused video. 
In addition, our method inherits several limitations of SVD. 
For example, we observed that in some cases fine-grained color details cannot be well reconstructed. %
This is mainly due the resolution of the VQ-VAE encoder, 
and since the starting frame is already encoded with artifacts, the generated video retains them.
Further, while SVD's generations suggest strong understanding of the physical world, there is still a lack of understanding regarding ``common sense'' and causal effect. For example, given an image of the famous moon landing, TRF generates a loop video in which the planted flag moves as if there was wind, which is not possible given the known context of the location. This is not only inaccurate, but could potentially bring ethical issues --e.g. the previous example could be misused as proof that the moon landing never happened. Video examples of these limitations are shown in our \projectpage.

Interestingly, there are some limitations of SVD that can be mitigated or resolved by our method. For example, SVD usually struggles with complex kinematic motions such as body limbs movement. Here, the generation tends to degrade throughout time, performing worse the further it is from the initial frame. On the other hand, TRF regularizes this through the bi-directional generation process, and can generate good-quality body motion between complex and distinct body poses. 

\section{Conclusion}
In this paper, we introduce bounded generation as a form of generalized control for pre-trained image-to-video models like SVD. We achieve so by proposing Time Reversal Fusion, a new sampling strategy that does not involve training or tuning of the original model, thereby preserving the model's inherent generalization capacity. We demonstrate bounded generation on three distinct settings that cover a diverse set of dynamics, and further curate a bounded generation dataset to show the effectiveness of TRF. We show that combining bounded generation and I2V models opens up opportunities for controlled video generation and provides a valuable avenue for probing the underlying dynamics within existing I2V models.

\clearpage  %

\bibliographystyle{splncs04}
\bibliography{main}
\end{document}